\documentclass{article}

\usepackage{arxiv}

\usepackage[utf8]{inputenc} 
\usepackage[T1]{fontenc}    
\usepackage{hyperref}       
\usepackage{url}            
\usepackage{booktabs}       
\usepackage{amsfonts}       
\usepackage{nicefrac}       
\usepackage{microtype}      
\usepackage{lipsum}		
\usepackage{graphicx}
\usepackage{natbib}
\usepackage{doi}

\usepackage{placeins}

\usepackage{listings}

\usepackage{amsmath}

\title{{\textbf{\MakeUppercase{Towards the Exploitation of LLM-based Chatbot for Providing Legal Support to Palestinian Cooperatives}}}}

\author{{\hspace{1mm}Rabee Al-Qasem} \\
    Data scientist\\
    Palestine\\
    \texttt{Rabee.qasem93@gmail.com} \\
    \And
    {\hspace{1mm}Banan Tantour} \\
    Legal Advisor \\
    Palestine\\
    \texttt{banan.tantour@hotmail.com} \\
    \And
    {\hspace{1mm}Mohammed Maree} \\
    Department of Information Technology,\\ Faculty of Engineering and Information Technology, \\Arab American Unievsrity, Jenin, Palestine \\
    \texttt{mohammed.maree@aaup.edu} \\
}

\renewcommand{\headeright}{}
\renewcommand{\undertitle}{}
\renewcommand{\shorttitle}{}

\hypersetup{
pdftitle={Towards the Exploitation of LLM-based Chatbot for Providing Legal Support to Palestinian Cooperatives},
pdfsubject={},
pdfauthor={Rabee Al-Qasem, Banan Tantour,Mohammed Maree},
pdfkeywords={Chatgpt, Artificial intelligence, Chatbots},
}

\begin{document}
\maketitle

\begin{abstract}
With the ever-increasing utilization of natural language processing (NLP), we started to witness over the past few years a significant transformation in our interaction with legal texts. This technology has advanced the analysis and enhanced the understanding of complex legal terminology and contexts. The development of recent large language models (LLMs), particularly ChatGPT, has also introduced a revolutionary contribution to the way that legal texts can be processed and comprehended. In this paper, we present our work on a cooperative-legal question-answering LLM-based chatbot, where we developed a set of legal questions about Palestinian cooperatives, associated with their regulations and compared the auto-generated answers by the chatbot to their correspondences that are designed by a legal expert. To evaluate the proposed chatbot, we have used 50 queries generated by the legal expert and compared the answers produced by the chart to their relevance judgments. Finding demonstrated that an overall accuracy rate of 82\%  has been achieved when answering the queries, while exhibiting an F1 score equivalent to 79\%.

\end{abstract}


\section{Introduction}
\label{sec:Introduction}

Natural Language Processing (NLP) has revolutionized the way we interact with legal texts. It has made it easier to analyze and comprehend complex legal texts \citep{becerra2018rise,dale2019law}. One of the most recent significant advancements in this field is the development of Large Language Models, and the development of chatbots that are based on such models \citep{brown2020language,rae2021scaling}, were ChatGPT is at the forefront of this development \citep{lee2022evaluating}. With its vast training data and powerful capabilities, ChatGPT has had a profound impact on global users. It provides them with intelligent conversational agents capable of understanding and responding to their queries. The integration of LLMs-powered chatbots has extended beyond the legal domain, finding applications in various fields. However, it is in the realm of legal discourse where these chatbots truly shine \citep{omar2023chatgpt}. They leverage their expertise to assist users in navigating complex legal terms and processes \citep{brooks2020artificial,fang2023ai,rajendra2022artificial}.

The huge improvement in LLM-based chatbot technology and the ease of integrating it seamlessly in the context of the legal domain has encouraged us to build a chatbot to provide answers to legal inquiries and questions about Palestinian cooperative law. We noticed that there have been numerous inquiries from cooperative societies and cooperative unions regarding private legal issues. This is mainly because the law is relatively new, having been issued at the end of 2017 \citep{cop_law} . Additionally, there is an urgent need to provide legal answers at all times, especially considering the need for a labor-intensive effort to answer such queries. Furthermore, considering the large number of cooperative members, which reached 58,883 at the end of 2021 as reported in \citep{cwa_report}, there is an urgent need for a chatbot that is available 24/7 to address their legal inquiries and provide timely assistance.
The rest of those article is organized as follows. In Section \ref{sec:literature}, we review the literature and discuss the related works. Section \ref{sec:dataset}, introduces the utilized dataset for testing and evaluating our proposed chatbot. In Section \ref{sec:method}, we discuss the proposed methodology. Section \ref{sec:results} present the experimental evaluation and results. In Section \ref{sec:Conclusion}, we conclude our work and point to the future directions of our research work in Section \ref{sec:futurework}.

\section{Literature review}
\label{sec:literature}

The use of machine learning (ML) techniques in the legal domain has long history with it a lot of research has integrated the two domains in many fields such as Legal document review \citep{mahoney2019framework,wei2018empirical}, Legal prediction \citep{sil2020novel}, Legal writing \citep{phelps2022alexa} , legal summarization \cite{elaraby2022arglegalsumm}, and Legal compliance \citep{mandal2017modular}. Prompting can be used to improve the performance of LLMs in different criteria and explore the effectiveness of using prompts in legal judgment prediction (LJP). \citep{trautmann2022legal} conduct experiments using data from the European Court of Human Rights and the Federal Supreme Court of Switzerland, comparing different prompts with multilingual language models (LLMs) such as mGPT, GPT-J-6B, and GPT-NeoX-20B. The results demonstrate that zero-shot prompt engineering can improve LJP performance with LLMs, yielding better macro-averaged F1 scores, precision, and recall compared to simple baselines. However, the performance of zero-shot learning still falls short of current supervised state-of-the-art results in the field. The paper also highlights the following key findings: prompting can enhance LLM performance in legal judgment prediction, multilingual LLMs can be effective even with training data in a single language, and while zero-shot learning holds promise, further improvements are needed to achieve state-of-the-art outcomes. The authors conclude by emphasizing the potential value of prompting for legal professionals and the accessibility benefits of multilingual LLMs in the field of legal natural language processing (NLP).  In an experiment, \citep{pettinato2023chatgpt} built a fictitious law professor who had a normal week of duties including teaching and community service planned out for her. then they used ChatGPT prompts for each task to test how well the system worked. For six of the seven tasks given, ChatGPT was able to produce workable first drafts in just 23 minutes. The most common tasks, including making a practice exam question or preparing a class handout, showed ChatGPT to be the most proficient. ChatGPT struggled with more complex tasks, especially those that had to do with education, although it still had the potential to save time in some cases. The experiment's findings indicate that ChatGPT, especially service-related jobs, has a lot of promise for reducing some components of the workload for law faculty. Additionally, ChatGPT may enable law professors to spend less time on specific teaching responsibilities, giving up more time for them to concentrate on pedagogy and create innovative teaching strategies. Finally, \citep{queudot2020improving} design and implementation of two immigration chatbots to advise their users about immigration legal questions and cases. One answers immigration-related questions, and the other answers legal questions from NBC employees. Both chatbots use supervised learning to learn embeddings for their answers. 

\section{Dataset}
\label{sec:dataset}
In our research work, we used 5 resources to acquire the input for our chatbot. To do this, we used three official documents which are Law No. 20 of 2017 on Cooperatives, Cooperatives Bylaws, and Housing Cooperatives Bylaws. Also, we created two sets of questions and answers datasets which we will discuss more in section \ref{sec:FLD}

\subsection{Formal Legal Documents }
\label{sec:FLD}

In order to give the Chatbot the legal context that it needs to answer legal questions, we needed to provide it with the legal documents that the lawyers and  Legal advisors depend on and use to answer legal questions but we made some reformatting for these documents where we only kept the necessary  articles and definitions these legal documents are: 

\begin{enumerate}
  \item  Law No. 20 of 2017 on Cooperatives: published in 2017 to govern the cooperative work in Palestine, under which the authority supervising the cooperative work sector in Palestine was established, known as the Cooperative Work Agency (CWA). It also specially regulates cooperative societies and unions.
cooperative members and the local community.
  \item Cooperatives Bylaws: It is the bylaws that govern the cooperative and the union, which regulate their work and the nature of their activity, based on the provisions of Law Decree No. 20 of 2017.
  \item Housing Cooperatives Bylaws: It is the bylaws that govern the housing cooperatives, which regulate their work and the nature of their activity, based on the provisions of Law Decree No. 20 of 2017.
\end{enumerate}

\subsection{Question and Answers Dataset }
\label{sec:QnA}

In order to support the chatbot to better understand the legal questions we created two other data set which contains a JSON file of questions about the Decree – Law No. (20) Of 2017 On Cooperative cooperatives the two data set are as following:

\subsubsection{Human Generated Question Answer Dataset}
We asked the legal advisor on cooperatives to create a dataset containing 40 questions and answers about different articles from the Decree – Law No. (20) Of 2017 On Cooperative. The questions and answers cover the basic topics of the definition of a cooperative, the requirements for forming a cooperative, the rights and responsibilities of cooperative members, and the role of the CWA in regulating cooperatives.

\subsubsection{Chatgpt Generated Question and Answers}
\begin{figure}[ht!]
  \caption{prompt  Structure}
  \centering
  \includegraphics[width=0.5\textwidth]{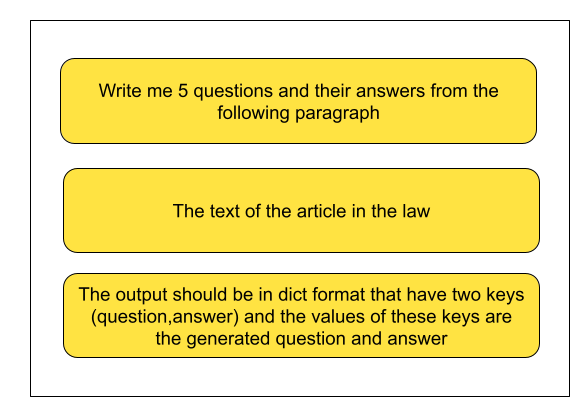}
  \label{fig:promt_strucutrer}
\end{figure}

We used the ChatGPT API to generate 5 questions and their corresponding answers for each article of Law No. (20) of 2017 on Cooperative. However, we needed to customize the answers to simulate the response of a real legal advisor. This involved starting the answer by referring to the article number in the law. To achieve this, we utilized the following prompt structure, as shown in Figure \ref{fig:promt_strucutrer}: we first requested the generation of the question and answer, then provided the article itself, and finally, to control the output, we asked ChatGPT to create a dictionary with two keys: "question" and "answer". After ChatGPT generated the dictionary, we appended it to another dictionary to collect the data. This process resulted in 350 questions and their corresponding answers.

This prompt helped us to control the output of the ChatGPT as we can see from the following code snippet

\begin{figure}[ht!]
  \centering
  \includegraphics[width=1 \textwidth]{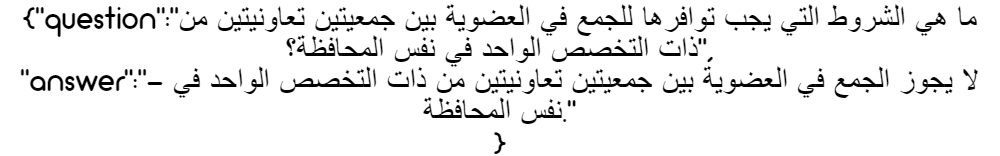}
\end{figure}

\section{Methodology}
\label{sec:method}

In our work, we encountered a vast amount of textual data that exceeded ChatGPT GPT-4's current processing limit at 8,192 tokens \citep{white2023chatgpt}. In order to take advantage of the ChatGPT API and overcome this obstacle, we utilized a comprehensive solution by employing LlamaIndex\citep{Liu_LlamaIndex_2022}. This proved to be a strategic decision as it enabled us to index large-scale datasets quickly and efficiently through its tailored features created specifically for Language Models (LLMs). The provided tools excel in generating vectors for every document while keeping them readily available.

To make the text data compatible with the ChatGPT API, we employed LlamaIndex to create an index encompassing all the legal documents and question-answer data at hand. Subsequently, we generated vectors for each document, ensuring that the input size did not exceed 8,192 tokens, while employing a chunk size of 600 tokens. The chosen chunk size of 600 tokens aligned with the requirements of the LLM. Moreover, we configured the maximum chunk overlap to be 50 tokens.

\begin{figure}[ht!]
  \caption{ChatGPT-based Chatbot workflow}
  \centering
  \includegraphics[width=0.5\textwidth]{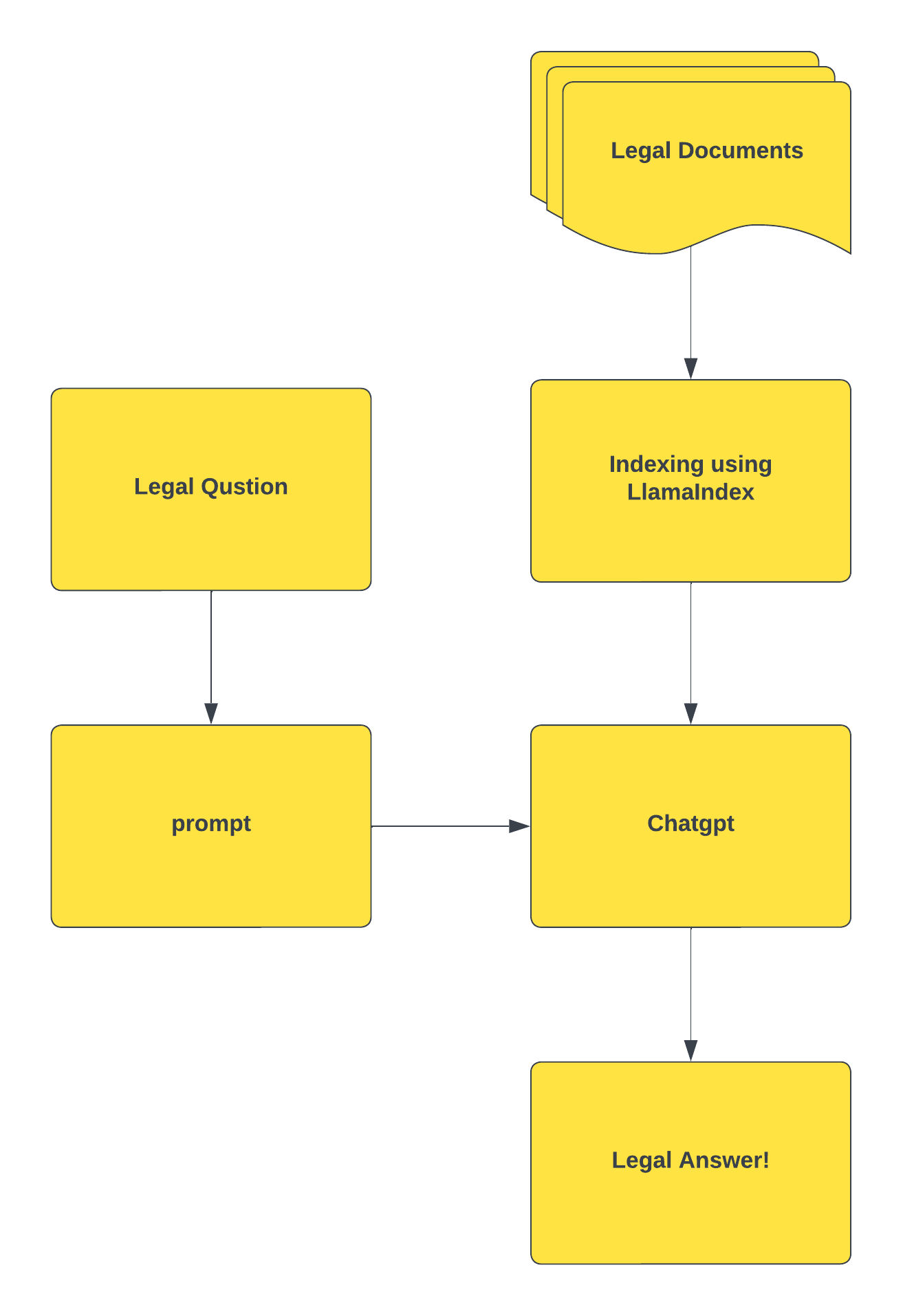}
  \label{fig:pipline}
\end{figure}

We efficiently stored the generated vectors within the index, enabling their swift retrieval whenever necessary. Leveraging the LlamaIndex query engine, which harnessed the power of ChatGPT in the background, we successfully addressed our legal queries and concerns. Figure \ref{fig:pipline} represents the comprehensive pipeline that we implemented for our case study, clearly demonstrating the use of LlamaIndex with ChatGPT and the subsequent vector generation and indexing of the legal documents.

\FloatBarrier

\section{Experimental Setup and Evaluation Results}
\label{sec:results}

After building the chatbot, we asked the legal advisor to write another 50 questions and their answers for testing purpuses. We then used the written questions to test the chatbot and compared the chatbot's answers to the legal advisor's answers. The chatbot was able to answer 41 questions in general. For example, we asked the chatbot about membership, financial statements, administrative issues, and how to register a new cooperative. The chatbot not only answered the right answers but in some cases, it also cited the law number and article (see Table \ref{tab:right_answers}.)

\begin{table}[ht!]
\caption{Right answers provided by the chatbot}
\centering
\includegraphics[width=0.95\textwidth]{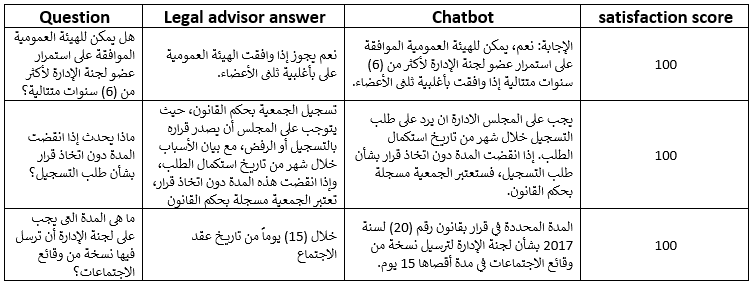}
\label{tab:right_answers}
\end{table}

\FloatBarrier

Although the chatbot was able to answer 41 questions correctly, not all of them were answered directly. Eight out of the 41 answers were relevant, but not direct. For example, when we asked the chatbot in Arabic "When does the management committee meet in the cooperative?", the chatbot gave us the correct answers, but it combined two answers. The first answer was the meeting of the management committee, and the second answer was the meeting of the general assembly. We have a lot of cases like this, and this is due to not having enough questions and answers for each article. Some articles of the law are short, so 5 questions and answers were enough to give the chatbot the context of the article. However, some articles are long and needed more questions and answers. 5 questions and answers were not enough to give the chatbot the context it needs to understand the difference, for example, between the meeting of the management committee and the general assembly. But in general, it gave us the right answer (See table \ref{tab:rel_answers}).

\begin{table}[ht!]
\caption{Related answers provided by the chatbot}
\centering
\includegraphics[width=0.95\textwidth]{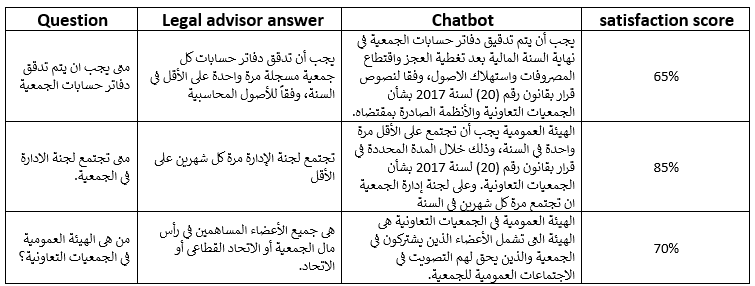}
\label{tab:rel_answers}
\end{table}

Finally, when we analyzed the rest of the wrong answers, we found that most of them were due to two reasons. First, there were not enough questions and answers for long articles, which required more explanation for the chatbot. Second, some articles had bylaws that needed to be provided to the chatbot. For example, when we asked the chatbot "Is it permissible to establish more than one general union?", which is illegal, the chatbot answered yes, which is a totally illegal act. (See table. \ref{tab:rong})

\begin{table}[ht!]
\caption{Wrong answers provided by the chatbot}
\centering
\includegraphics[width=0.95\textwidth]{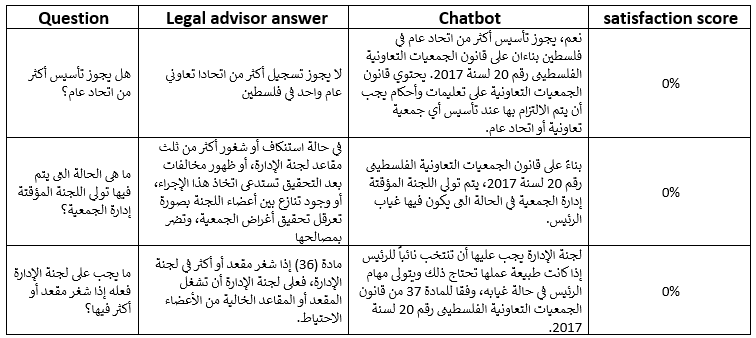}
\label{tab:rong}
\end{table}

\FloatBarrier

To measure the performance of our chatbot, we used the following metrics:

\textbf{Overall accuracy:} This metric is calculated by dividing the total number of correct answers by the total number of questions asked. The equation for overall accuracy is:

\begin{equation}
\text{Overall accuracy} = \frac{\text{Total number of correct answers}}{\text{Total number of questions}} \times 100
\end{equation}

In this case, the chatbot achieved an overall accuracy of 41/50, or 82\%. This is a good result, as it means that the model was able to correctly answer 82\% of the questions asked. and if we provided more data for it it will increase its accuracy 

\textbf{Overall satisfaction:}  Many studies used satisfaction scores with other metrics to evaluate their trained chatbots \citep{casas2020trends}, but in our case, we used only the satisfaction score. We did this by letting the legal counsel give a mark for how satisfied they were with the answer. For example, in the case of a right answer, the legal counsel was very satisfied, so she gave it a score of 100\%. For wrong answers, the score was 0\%. For related answers, the score was between 60\% and 85\%. We calculated this by computing the average of all satisfaction scores and the total number of questions.

\begin{equation}
\text{Average satisfaction score} = \frac{\sum_{i=1}^n S_i}{n}
\end{equation}

where Si is the satisfaction score for the $i$th question and n is the total number of questions.

In this case, the chatbot achieved an average satisfaction 78.3\%. Which is also a good result for our chatbot.

\textbf{Confusion matrix:} To measure the performance of the chatbot, we used the precision, recall, and F1 score. The confusion matrix is used to measure classification models and use the actual value and the predicted value of the model to compute the precision and recall and then the F1 score. However, since we didn't train the chatbot, we assumed that all the answers of the legal counselor are correct. This assumption will affect the precision value, as the chatbot will not be penalized for incorrectly identifying an answer as wrong.

\begin{enumerate}

\item \textbf{Precision:} Since we made the assumption that there are no wrong actual answers the precision for class  (wrong) is 0,  and the  precision for class 1 (right + related ) is 1.0, indicating that our chatbot correctly predicted all instances as "right" or "related."

\item \textbf{Recall:} The recall for class  (wrong) is 0, indicating that our chatbot did not correctly identify any instances as "wrong." The recall for class  (right/related) is 0.79, meaning that our chatbot correctly identified 79\% of instances labeled as "right" or "related."

\item \textbf{F1-score}: The F1-score for class  (wrong) is 0, which aligns with the precision and recall being 0. The F1-score for class  (right/related) is 0.88, indicating a relatively good balance between precision and recall for this class.

\item The \textbf{accuracy} of our chatbot is reported as 0.79, meaning it correctly predicted the label for 79\% of the instances in the questions that we provided.

For more information and details on the developed chatbot, please refer to our GitHub Repository at the following link: \ \href{https://github.com/rabeeqasem/llm_chatbot_legal}{Github}

\end{enumerate}

\section{Conclusion}
\label{sec:Conclusion}

In this paper, we introduced our LLM-based legal chatobot that aims to assist Palestinian cooperatives and their members in finding relevant answers to their legal inquiries. Our objective was to provide accurate and reliable support 24/7 by leveraging the vast amount of publicly-available legal documents that we were able to acquire. After utilizing the chatbot on this extensive dataset, we achieved an overall accuracy  of 82\% and an F1 score of 79\%.

However, as we encountered an enormous volume of text data, we faced challenges with the chatbot's processing limit in terms of the maximum amount of text data that can submitted to through ChatGPT's API. To overcome this obstacle, we implemented a technique called 'vectorization' using LlamaIndex. This process converted the text data into a format that the chatbot could effectively utilize.

Whilst serving as a valuable legal aid for diverse cooperative members, there are certain limitations inherent to the chatbot that should be acknowledged. Our study uncovered instances where the chatbot provided incorrect answers, which could potentially lead users to unintentionally violate legal regulations. Consequently, we believe that continuous development and improvement of the chatbot are necessary to enhance its accuracy and reliability.
Furthermore, it is crucial to be transparent about the chatbot's limitations and ensure that users have access to comprehensive information. This will enable them to make informed decisions about utilizing the chatbot's services. By refining the chatbot and openly communicating its capabilities, we can harness its potential as an invaluable tool for delivering reliable legal support to a diverse audience.

\section{Future Works}
\label{sec:futurework}

In the future work, we plan to address the challenges highlighted in the previous section through the following steps. First, we will focus on increasing the size of the used dataset by formalizing additional question, with their relevance judgments, and also ensuring that the expert questions are close in terms of their number to those generated by the chatbot. Second, we plan to post-process the answers produced bu the chatbot to further enhance and improve the overall quality, i.e. accuracy of the answers. This may also require the exploitation of legal domain knowledge and semantic resources that can be further utilized for reformulating users' questions in a more legal relevant context.

\bibliographystyle{apalike}
\bibliography{references}

\begin{thebibliography}{}
\bibitem[Becerra, 2018]{becerra2018rise}
Becerra, S.~D. (2018).
\newblock The rise of artificial intelligence in the legal field: where we are
  and where we are going.
\newblock {\em J. Bus. Entrepreneurship \& L.}, 11:27.
\end{thebibliography}

\end{document}